\def\year{2020}\relax
\documentclass[letterpaper]{article}
\usepackage[hyphens]{url}  
\usepackage{graphicx} 
\usepackage{graphicx}  
\usepackage{aaai20}
\usepackage{times}
\usepackage{helvet}
\usepackage{courier}
\usepackage[semicolon]{natbib}
\usepackage[euler]{textgreek}
\usepackage{amssymb}
\usepackage{amsmath}
\usepackage{booktabs}

\usepackage{array}
\usepackage{tabularx}
\usepackage{multirow}
\usepackage{blindtext}
\usepackage{subfig}
\usepackage[super]{nth}
\usepackage{textcomp}
\renewcommand{\textrightarrow}{$\rightarrow$}
\title{Guided Weak Supervision for Action Recognition with Scarce Data to Assess Skills of Children with Autism}
\author{
Prashant Pandey,\textsuperscript{\rm 1}
Prathosh AP,\textsuperscript{\rm 1}
Manu Kohli,\textsuperscript{\rm 1}
Josh Pritchard\textsuperscript{\rm 2}\\
\textsuperscript{\rm 1}IIT Delhi,
\textsuperscript{\rm 2}Florida Institute of Technology\\
\{bsz178495, prathoshap\}@iitd.ac.in, manu.kohli@cogniable.tech,
josh@factari.com
}
\begin{document}
\maketitle
\begin{abstract}
Diagnostic and intervention methodologies for skill assessment of autism typically requires a clinician repetitively initiating several stimuli and recording the child's response. In this paper, we propose to automate the response measurement through video recording of the scene following the use of Deep Neural models for human action recognition from videos. However, supervised learning of neural networks demand large amounts of annotated data that is hard to come by. This issue is addressed by leveraging the `similarities' between the action categories in publicly available large-scale video action (source) datasets and the dataset of interest. A technique called Guided Weak Supervision is proposed, where every class in the target data is matched to a class in the source data using the principle of posterior likelihood maximization. Subsequently, classifier on the target data is re-trained by augmenting samples from the matched source classes, along with a new loss encouraging inter-class separability. The proposed method is evaluated on two skill assessment autism datasets, SSBD \citep{Rajagopalan_2013_ICCV_Workshops} and a real world Autism dataset comprising 37 children of different ages and ethnicity who are diagnosed with autism. Our proposed method is found to improve the performance of the state-of-the-art multi-class human action recognition models in-spite of supervision with scarce data.
\end{abstract}
\section{Introduction}
\begin{figure}[hbt!]
    \begin{minipage}[t]{.112\textwidth}
        \centering
        \includegraphics[width=\textwidth,height=0.512\textwidth]{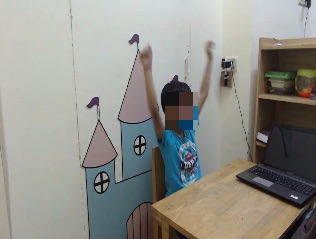}
        a) RGB frame from Autism data.
        \label{fig:1}
    \end{minipage}
    \begin{minipage}[t]{.115\textwidth}
        \centering
        \includegraphics[width=\textwidth,height=0.5\textwidth]{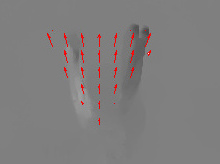}
        b) Optical flow for (a).\label{fig:2}
    \end{minipage}
    \begin{minipage}[t]{.114\textwidth}
        \centering
        \includegraphics[width=\textwidth,height=0.506\textwidth]{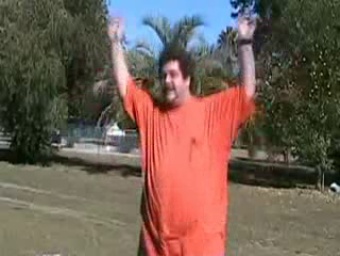}
        c) RGB frame from Kinetics data.\label{fig:3}
    \end{minipage} 
    \begin{minipage}[t]{.114\textwidth}
        \centering
        \includegraphics[width=\textwidth,height=0.509\textwidth]{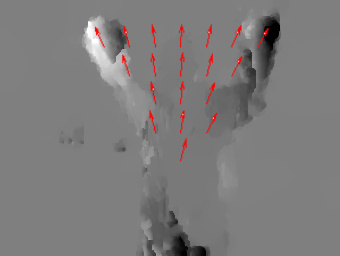}
        d) Optical flow for (c).\label{fig:2}
    \end{minipage}
    \\\\\\
    \begin{minipage}[t]{.114\textwidth}
        \centering
        \includegraphics[width=\textwidth,height=0.503\textwidth]{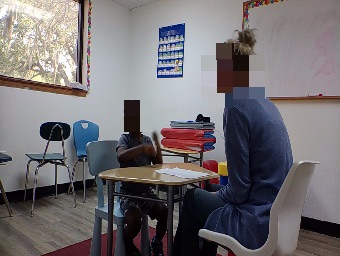}
        e) RGB frame from Autism data.\label{fig:1}
    \end{minipage}
    \begin{minipage}[t]{.114\textwidth}
        \centering
        \includegraphics[width=\textwidth,height=0.506\textwidth]{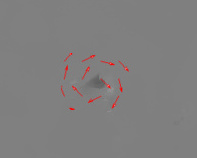}
        f) Optical flow for (e).\label{fig:2}
    \end{minipage}
    \begin{minipage}[t]{.114\textwidth}
        \centering
        \includegraphics[width=\textwidth,height=0.505\textwidth]{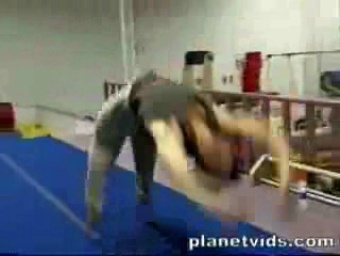}
        g) RGB frame from HMDB51.\label{fig:3}
    \end{minipage} 
    \begin{minipage}[t]{.114\textwidth}
        \centering
        \includegraphics[width=\textwidth,height=0.505\textwidth]{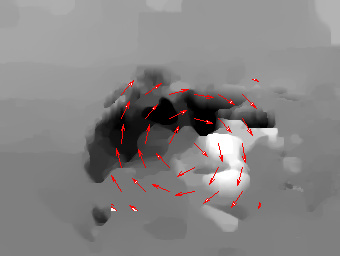}
        i) Optical flow for (g).\label{fig:2}
    \end{minipage}
    \caption{Illustration of the similarity between the action classes in two different datasets in the optical flow domain. One representative RGB and Flow frame is depicted in each case: The directional closeness of the optical flow frames can be observed despite being from unrelated classes. (a)-(b) `Arms up' action in Autism data is close to (c)-(d) `Jumping jacks' action in Kinetics \citep{kay2017kinetics} data. (e)-(f) `Rolly polly' action in Autism data is close to (g)-(h) `Flic flac' action in HMDB51 \citep{kuehne2011hmdb} data.}
    \label{fig:sim-class}
\end{figure}
Autism is a complex neuro-developmental disorder that manifests in children during preschool years \citep{rapin1991autistic,CDC} as deficits in communication, social skills and stereotypical repetitive behavior. In the last two decades, the prevalence rate of Autism Spectrum Disorder (ASD) has grown by more than 150\% \citep{CDC}. 
It is well-established that early intervention services modelled on behavior therapies yield the best outcomes for children diagnosed with autism \citep{estes2015long} leading to significant societal benefits and cost savings.
However, resource and expert scarcity in low resource settings result in delay in the initiation of the treatment process due to lack of identification of the disorder. Families have limited access to diagnostic and evidence-based treatment options because of affordability, lack of insurance support and non-availability of physical infrastructure.
In the developing countries, there are entire swaths of children in dire need of treatment who remain untreated and unseen.
Much of this strain and inaccessibility could be alleviated by incorporating technology to assist in early screening and automate the assessment and initial treatment planning process, which is a prerequisite to deliver individualized behavior treatment to children. 
Skill assessment processes for autism typically involves invoking instructions to a child, monitoring and recording their responses as they occur. This requires a trained clinician to engage with the child, perform mundane repetitive tasks such as recording the child's observation and human action responses to fixed set of stimuli. With the advent of tremendously powerful modern Deep learning techniques, one can hope to automate a lot of such tasks bringing affordability and improved access in value chain of autism screening, diagnosis and behavioral treatment activities while reducing the dependence on trained clinicians.  Specifically, in this paper, we examine the application of human action recognition from video recordings for tracking the physical 
behavior of children diagnosed with ASD or otherwise to build cognitive and functional assessments. 
\section{Motivation}
Human action recognition is typically solved in a supervised machine learning setting.  Most of the successful models employ Convolutional Neural Networks (CNN) as their backbone. Two stream networks \citep{simonyan2014two}, 3-Dimensional (3D) Convolutional Neural Networks \citep{tran2015learning} and Convolutional long short-term memory (LSTM) networks \citep{donahue2015long} are some of the state-of-the-art action recognition approaches. 
Inflated 3D CNN \citep{carreira2017quo} or I3D is found to be one of the top performers on the standard benchmarks like UCF101 \citep{soomro2012ucf101} and HMDB51 \citep{kuehne2011hmdb} datasets. Temporal Segment Networks \citep{wang2016temporal} or TSN is another example of two stream networks which has better ability to model long range temporal dependencies.

The objective of this work is to apply human action recognition algorithms to  evaluate the responses of the children with autism, to measure stimulus in the area of imitation, listener response and motor skills. To accomplish this using already established methods require large amounts of expert annotated data corresponding to the particular classes of actions to be recognized. The process of data collection and annotation  is non-trivial because of non-availability of expert annotators, lack of co-operation from children and is also very time consuming. 
Despite these limitations, there is abundant availability of  large-scale  public datasets that contain thousands of annotated video clips corresponding to hundreds of action classes. Further, human action classes share a lot of similarity (in a well-defined feature space like optical flow \citep{horn1981determining}) in-spite of being disjoint as shown in Figure \ref{fig:sim-class}.
For example, intuitively, the action classes `playing piano' and `typing the keyboard' can be thought similar in a suitable feature space. Motivated by these observations,  in this paper we address the following question - \textit{Given a target data distribution with few annotated samples, can the availability of a large-scale annotated source dataset with `similar' attributes as the target data, be leveraged, to improve the generalization abilities of classifiers trained on target data?} Specifically, we attempt to weakly supervise the task of human action recognition in a guided way using large-scale publicly available datasets. 
\section{Related Work}
State-of-the-art action recognition models \citep{zolfaghari2018eco, lin2019tsm, ghadiyaram2019large, carreira2017quo, wang2016temporal} are deep neural networks which overfits easily to the dataset with fewer samples leading to poor generalization. 
Few-Shot Learning (FSL) action recognition methods \citep{Yang_2018_CVPR, Zhu_2018_ECCV, Wang_2018_CVPR, Mandal_2019_CVPR} are characterised by different labels between source and target but a similar feature space. FSL evaluates on novel classes with limited training examples but these novel classes are sampled from the same dataset. The large scale public datasets are not good candidate for source in FSL as there is significant domain shift that exists between the source and target data. 
\section{Proposed Method}
For a given video, there exists transformations such as optical flow, that are non-unique mappings of the video space. This suggests that given multiple disjoint set of action classes, there  can be spaces (such as flow) where a given pair of action classes may lie `close' albeit they represent different semantics in the RGB-space. For example, the optical flow characteristics of a `baseball-strike' class and `cricket-slog' class can be imagined to be close.  Further, there exists large-scale public datasets (e.g., Kinetics \citep{kay2017kinetics}) that encompasses a large number of annotated videos for several action classes. Thus, if one can find the classes in the public datasets that are `close'  to a given class in the dataset of interest, then the videos from the public dataset can be  potentially used for augmentation resulting in regularization. In the subsequent sub-sections, we will formalize the aforementioned idea and describe a procedure to find the closer classes and use it for data augmentation.
Let $\mathcal{X}$ denote the sample space encompassing the elements of transformed videos (e.g., optical flow). Let $\mathbf{P^s(x_s)}$ and $\mathbf{P^t(x_t)}$ be two distributions on $\mathcal{X}$ called the source and target distributions respectively. Suppose a semantic labeling scheme is defined both on $\mathbf{P^s(x_s)}$ and $\mathbf{P^t(x_t)}$. That is, let $\mathcal{Y}_s=\{ \mathbf{y_s^1}, \mathbf{y_s^2}..., \mathbf{y_s^N} \}$ and $\mathcal{Y}_t=\{\mathbf{y_t^1}, \mathbf{y_t^2}..., \mathbf{y_t^M}\}$ be the source and target class labels that are assigned for the samples of $\mathbf{P^s(x_s)}$ and $\mathbf{P^t(x_t)}$ respectively which in-turn defines the joint distributions $\mathbf{P^s(x_s,y_s)}$ and $\mathbf{P^t(x_t,y_t)}$. $\mathbf{N}$ and $\mathbf{M}$ are the respective number of source and target classes. Let $\mathcal{D}_s= \{\mathbf{(x_{s}, {y_{s}})} \}$ and $\mathcal{D}_t= \{\mathbf{(x_{t}, {y_{t}})} \}$ denote the tuples of samples drawn from the two joint distributions $\mathbf{P^s}$ and $\mathbf{P^t}$, respectively. Suppose a parametric discriminative classifier (Deep Neural Network) is learned using  $\mathcal{D}_t$  to obtain estimate of the conditional distribution $\mathbf{P_\theta^t(y_t|x_t)}$ where $\theta$ denotes the parameters of the neural network. 
With these notations, we consider the case where the cardinality of $\mathcal{D}_t$ is much less than that of $\mathcal{D}_s$ implying that the amount of supervised data in the case of target distribution is much less than that of the source distribution. In such a case, $\mathbf{P_\theta^t(y_t|x_t)}$ trained on $\mathcal{D}_t$ is deemed to overfit and hence doesn't generalize well. If there exists a $\mathbf{y_s^p} \in \mathcal{Y}_s$ that is `close' to $\mathbf{y_t^q} \in \mathcal{Y}_t$, then samples drawn from  $\mathbf{P^s(x_s|y_s=y_s^p)}$ can be used to augment the class $\mathbf{y_t^q}$ for  re-training the model $\mathbf{P_\theta^t(y_t|x_t)}$. In the subsequent sub-section, we describe a procedure to find the `closest' $\mathbf{y_s^p} \in \mathcal{Y}_s$, given $\mathbf{y_t^q} \in \mathcal{Y}_t$ and a model $\mathbf{P_\theta^t(y_t|x_t)}$ trained on $\mathcal{D}_t$.  
\subsection{Guided Weak Supervision}
\begin{figure*}
\centering
\scalebox{0.94}{
  \includegraphics[width=\textwidth]{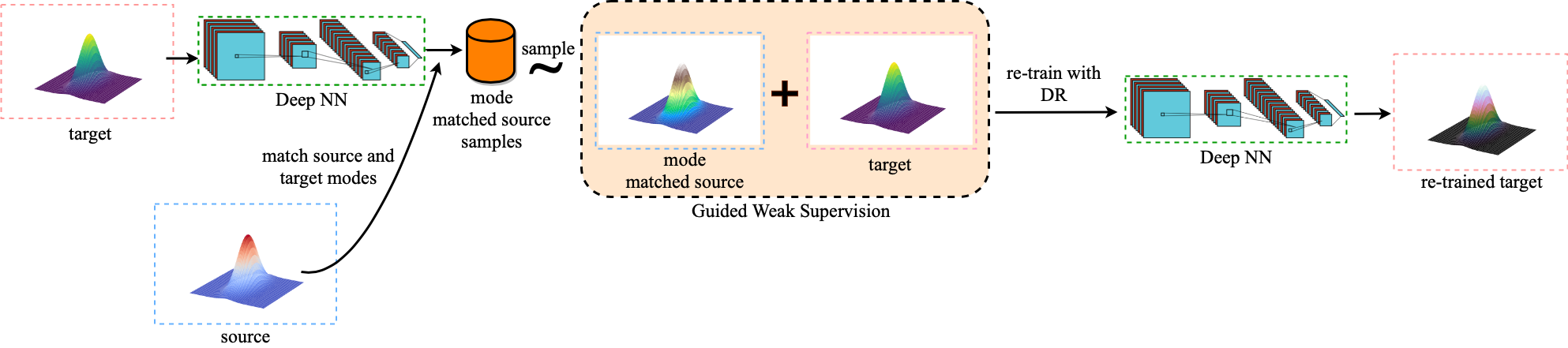}
  }
  \caption{The overall procedure of the proposed method. A classifier trained on the target data is used to match the modes (classes) of the source data. Target classes are augmented with samples from the matched source classes (Guided Weak Supervision). Finally, the classifier is re-trained with the augmented target classes along with Directional Regularization loss.}
  \label{fig:overall_flow}
\end{figure*}
Videos lie in a very high dimensional space and are of variable length in general. Thus, standard vector distance metrics are not feasible to measure the closeness of two video objects. Further, the objective here is to quantify the distance between the classes as perceived by the discriminative model (classifier)    $\mathbf{P_\theta^t(y_t|x_t)}$ so that the data augmentation is sensible. Thus, we propose to use the maximum posterior likelihood principle to define the closeness between two classes. 
Let $\mathcal{X}_{(\mathbf{y_s=y_s^p})}=\mathbf{\{x_{s1},x_{s2},....,x_{sl}\}}$ denote the samples drawn from $\mathbf{P^s(x_s|y_s=y_s^p)}$.  Now  $\mathbf{P_\theta^t(y_{tj}|x=x_{sj})}$,  $\mathbf{j \in \{1,2..,l\}}$ denotes the posterior distribution of the target classes  $\mathcal{Y}_t$ given the $\mathbf{j^{th}}$ feature vector from the source class $\mathcal{X}_{(\mathbf{y_s=y_s^p})}$. With this, a joint posterior likelihood $\mathcal{L}_\mathbf{y_t|x_s}$ of a class $\mathbf{y_s^p}$ can be defined as observing the target classes given a set of features $\mathcal{X}_{(\mathbf{y_s=y_s^p})}$ drawn from a particular source class $\mathbf{y_s^p}$. Mathematically,
\begin{equation}
\mathcal{L}_\mathbf{y_t|x_s} =  \mathbf{P_\theta^t(y_{t1},y_{t2},....,y_{tl}|x_{s1},x_{s2},....,x_{sl}})
\end{equation} where $\mathbf{x_{sj}}$, $\mathbf{j \in \{1,2..,l\}}$, are from the class $\mathbf{y_s^p}$. If it is assumed that $\mathbf{x_{sj}}$ are drawn i.i.d., one can express Eq. 1 as, 
\begin{equation}\mathcal{L}_\mathbf{y_t|x_s}  = \mathbf{\prod_{j=1}^l} \mathbf{P_\theta^t(y_{tj}|x_{sj}})
\end{equation} 
The parameters $\theta$ of the discriminative model created using $\mathcal{D}_t$ are independent of $\mathcal{X}_\mathbf{(y_s=y_s^p)}$ and are fixed during the evaluation of $\mathcal{L}_\mathbf{y_t|x_s}$, which implies that $\mathbf{y_{ti}|x_{si}}$ is independent of $\mathbf{x_{si}}$ $\mathbf{\forall  i \neq j}$ thus leading to Eq. 2. The posterior likelihood in Eq. 2 can be evaluated for every target class $\mathbf{y_t=y_t^q}$, $\mathbf{q \in \{1,2,...,M\}}$, denoted by $\mathcal{L}_\mathbf{y_t=y_t^q|x_s}$ called the target-class posterior likelihood corresponding to  the features from source class  $\mathbf{y_s^p}$ under the learned target classifier $\mathbf{P_\theta^t(y_t|x_t)}$. Mathematically,
\begin{equation}
\mathcal{L}_\mathbf{y_t=y_t^q|x_s} = \mathbf{\prod_{j=1}^l} \mathbf{P_\theta^t(y_{tj}=y_t^q|x_{sj})}
\end{equation} 
With this definition of the target-class posterior likelihood, we define the matched source class $\mathbf{y_s^*|y_t^q} \in \mathcal{Y}_s$ to a given target class $\mathbf{y_t^q}$ as follows:
\begin{equation}
\mathbf{y_s^*|y_t^q} =\underset{\mathcal{Y}_s}{\operatorname{argmax}} \; \mathcal{L}_{\mathbf{y_{t}=y_t^q|x_s}}
\end{equation} 
Note that the definition of $\mathcal{L}_{\mathbf{y_t=y_t^q|x_s}}$ is specific to a source-target class pair and therefore all $\mathbf{x_{sj}}$ in the objective function of the optimization problem in Eq. 4 comes from a particular source class. Thus, one can employ the discriminative classifier trained on the target data to find out the `closest' matching source class as the one that maximizes the posterior likelihood of observing that class as the given target class under the classifier. Every class in the joint distribution can be looked as a `mode' and the goal here is to match the classes (`modes') in the joint distribution of the source and target distributions. Figure \ref{fig:sim-class} demonstrates the idea of mode matching through examples. Optical flow frames of the target and the source classes have similar visual properties indicating the closeness. 
Once the matched source class is determined for every given target class, a  set of source classes matched is defined as $\mathcal{Y}_s^* = \{ \mathbf{y_s^{*1}, y_s^{*2},..,y_s^{*M} } \}$.  
Now, the discriminative classifier $\mathbf{P_\theta^t}$ can be re-trained on the samples from the source dataset corresponding to $\mathcal{Y}_s^*$  in a supervised way with class labels being the corresponding  $\mathbf{y_t^q}$ for every $\mathbf{y_s^{*}}$. This procedure thus increases the quantity and variety of the training data for $\mathbf{P_\theta^t}$ and we call it as Guided Weak Supervision (GWS).
 \subsection{Directional Regularization}
 The procedure of GWS described in the previous sub-section effectively changes the semantic meaning of the matched source classes to the semantic meaning of the target classes. Thus, it is possible to train a classifier on the source data to discriminate between the matched source classes $\mathcal{Y}_s^* = \{ \mathbf{y_s^{*1}, y_s^{*2},..,y_s^{*M} } \}$. Suppose such a classifier is denoted by $\mathbf{P_\phi^s(y_s^*|x_s)}$, where $\phi$ are the model parameters. We assume that $\mathbf{P_\phi^s(y_s^*|x_s)}$ and $\mathbf{P_\theta^t(y_t|x_t)}$ have the same architectural properties. Also, it is assumed that the source dataset is larger in size and more diverse as compared to the target dataset.
 This implies that $\mathbf{P_\phi^s(y_s^*|x_s)}$  has better generalization abilities compared to $\mathbf{P_\theta^t(y_t|x_t)}$ (This fact is corroborated empirically in the experiment section). We propose to leverage this fact in improving the generalization capabilities of $\mathbf{P_\theta^t(y_t|x_t)}$ using $\mathbf{P_\phi^s(y_s^*|x_s)}$. Further, during the training of $\mathbf{P_\phi^s(y_s^*|x_s)}$ with samples from $\mathcal{Y}_s^*$, it is desirable that the separation that is achieved between the classes in $\mathcal{Y}_s^*$ under the classifier $\mathbf{P_\phi^s(y_s^*|x_s)}$ is `preserved' during the training of  $\mathbf{P_\theta^t(y_t|x_t)}$ with samples from $\mathcal{Y}_s^*$. We propose to accomplish the aforementioned properties by imposing a regularization term during the training of $\mathbf{P_\theta^t(y_t|x_t)}$. Specifically, we propose to push the significant directions of the parameter matrix of $\theta$  towards that of the parameter matrix of $\phi$. Note that  $\phi$ is fixed during the training of $\mathbf{P_\theta^t(y_t|x_t)}$. Intuitively, this implies that the significant directions of the target parameters should follow that of the source parameters.
 Mathematically, let $\mathbf{M}_\theta$ and $\mathbf{M}_\phi$ be two square matrices formed by re-shaping (without any preference to particular dimensions) the parameters $\theta$ and $\phi$, respectively. We perform an Eigenvalue Decomposition on $\mathbf{M}_\theta$ and $\mathbf{M}_\phi$ to obtain matrices of eigenvectors $\mathbf{E}_\theta$ and $\mathbf{E}_\phi$, respectively. Let $\mathbf{\hat{E}}$ denote the truncated versions of $\mathbf{E}$ with first $k$ significant (a model hyperparameter) eigenvectors. Under this setting, we desire the significant directions $\mathbf{\hat{E}}_\theta$ and $\mathbf{\hat{E}}_\phi$ to be aligned. Mathematically, if they are perfectly aligned, then 
\begin{equation}
{\mathbf{\hat{E}}_\theta}^{\mathbf{T}}\mathbf{\hat{E}}_\phi = \mathbf{I}_k
\end{equation} where $\mathbf{I}_k$ is a $k$-dimensional identity matrix and $\mathbf{T}$ denotes the transpose operation. Thus, any deviation from the condition laid in Eq. 5 is penalized by minimizing the Frobenius norm of the deviation. We term this as Directional Regularization (DR) denoted as $\mathrm{L}_{DR}$ given by the following equation:
\begin{equation}
\mathrm{L}_{DR}= \|{\mathbf{\hat{E}}_\theta}^{\mathbf{T}}\mathbf{\hat{E}}_\phi -\mathbf{I}_k\|_{F}
\end{equation} 
where $|.|_F$ denotes the Frobenius norm of a matrix. Note that this regularizer imposed on $\theta$ during the training of $\mathbf{P_\theta^t(y_t|x_t)}$ ensures that the directions of separating hyperplanes of the target classifier is encouraged to follow those of the source classifier trained with the matched classes. Thus the final objective function during the re-training of the target classifier is as follows:
\begin{equation}
\mathrm{L}_{Total}= -\mathbf{\sum_{j=1}^{l}}\mathbf{y_{tj}}\log \mathbf{\hat{y}{_{tj}}} + \|{\mathbf{\hat{E}}_\theta}^{\mathbf{T}}\mathbf{\hat{E}}_\phi -\mathbf{I}_k\|_{F}
\end{equation} 
where $\mathbf{\hat{y}{_{tj}}}$ is the predicted target class. Thus in summary, given a small amount of data from the target distribution, the proposed method (1) trains a classifier on the target samples, (2) determines the closest classes from the source distribution to all the target classes using the target classifier, (3) trains a new (relatively robust) classifier on the samples from the source distribution with re-labeled source classes (matched with the target classes), (4) uses the samples of the matched source classes to re-train the target classifier along with Directional Regularization that builds the final model for the target data. This entire procedure is pictorially depicted in Figure \ref{fig:overall_flow}.
\subsection{Implementation}
The idea of GWS detailed in the previous section assumes that the source and the target distributions are `similar' under a certain feature transformation. The raw video data in the RGB space does not adhere to such assumptions because of the variability of the content/scene/subjects etc. However, transformations such as optical flow masks most of the non-motion related information and exaggerates the motion information. In this domain, it is reasonable to assume that the source and target action classes are `similar' (Refer Figure \ref{fig:sim-class}). We match each target Autism class to a source class using the baseline Autism model $\mathbf{P_\theta^t(y_t|x_t)}$. Since a probabilistic Softmax layer is used at the output of the classifier, the matched source class, for every given target class, can be simply taken to be that source class whose samples get labeled as the given target class most of the times as compared to all other source classes. The optimization problem in Eq. 4 can be approximated as follows:
\begin{flalign}
\mathbf{y_s^*|y_t^q} &=\underset{\mathcal{Y}_s}{\operatorname{argmax}} \; \mathcal{L}_{\mathbf{y_{t}=y_t^q|x_s}} \\
&= \mathbf{\prod_{j=1}^l} \mathbf{P_\theta^t(y_{tj}=y_t^q|x_{sj})}\\
& \approx \underset{\mathcal{Y}_s}{\operatorname{argmax}} \Bigg[ \underset{\mathbf{j\in \{1,..l\}}}{\operatorname{Count}} \bigg( \mathbf{y_t^q}= \underset{\mathcal{Y}_t}{\operatorname{argmax}} \; \mathbf{P_\theta^t(y_{tj}|x_{sj}}) \bigg) \Bigg]
\end{flalign} 
We propose to use state-of-the-art action recognition models such as I3D and TSN that employ optical flow streams. Baseline target Autism model $\mathbf{P_\theta^t(y_t|x_t)}$ is obtained by training Autism data using an I3D or TSN architecture by initializing their weights  with pre-trained Kinectics/ImageNet/HMDB51 models. GWS and re-training with Directional Regularization are performed only on the flow stream without altering the RGB stream. Once the source classes are matched, the samples from the matched source classes are trimmed to a fixed length of average duration of the target clips ($\approx$ 2 seconds in the Autism data). This is accomplished by trimming each video sample of the matched source classes using an overlapping window approach and taking that portion of the  video which has the highest Softmax prediction score under the baseline target classifier for that particular target class.  These trimmed (or action localized) video clips from the source data are used for data augmentation to re-train the target classifier with the Directional Regularization loss. Both the RGB and re-trained optical flow streams are combined for final prediction. All the data is pre-processed to generate the optical flow frames using the TVL1 algorithm \citep{zach2007duality}. The code associated with this paper can be found at \url{https://github.com/prinshul/GWSDR}.
\begin{figure*}[hbt!]
\captionsetup{justification=raggedright}
    \begin{minipage}[t]{.2471\linewidth}
        \centering
        \includegraphics[width=1.0\linewidth,height=0.4\linewidth]{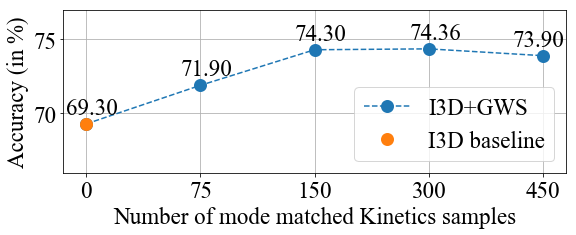}
        a) GWS on I3D using Kinetics samples.\label{fig:i3ddm}
    \end{minipage}
    \begin{minipage}[t]{.2471\linewidth}
        \centering
        \includegraphics[width=1.0\linewidth,height=0.4\linewidth]{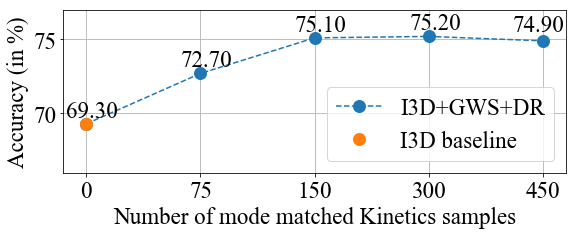}
        b) GWS+DR on I3D using Kinetics samples.\label{fig:i3ddr}
    \end{minipage}
    \begin{minipage}[t]{.2471\linewidth}
        \centering
        \includegraphics[width=1.0\linewidth,height=0.4\linewidth]{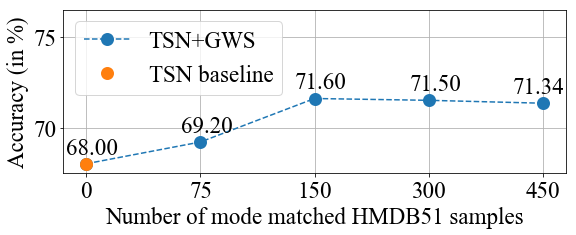}
        c) GWS on TSN with HMDB51 samples.\label{fig:tsndm}
    \end{minipage}
    \begin{minipage}[t]{.2471\linewidth}
        \centering
        \includegraphics[width=1.0\linewidth,height=0.4\linewidth]{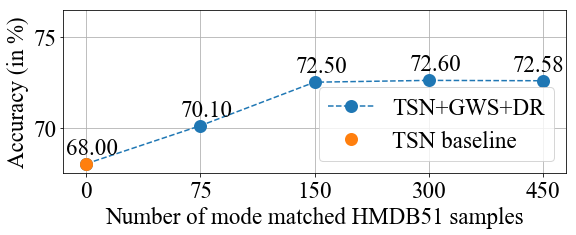}
        d) GWS+DR on TSN with samples from HMDB51.\label{fig:tsndr}
    \end{minipage}
    \caption{Performance on I3D and TSN with different baselines and different amount of augmented source (Kinetics and HMDB51) samples using  GWS and DR.}
    \label{fig:i3dTSN}
\end{figure*}
\section{Autism Dataset}
\begin{figure}[hbt!]
\captionsetup{justification=raggedright}
    \begin{minipage}[t]{.152\textwidth}
        \centering
        \includegraphics[width=\textwidth,height=0.44\textwidth]{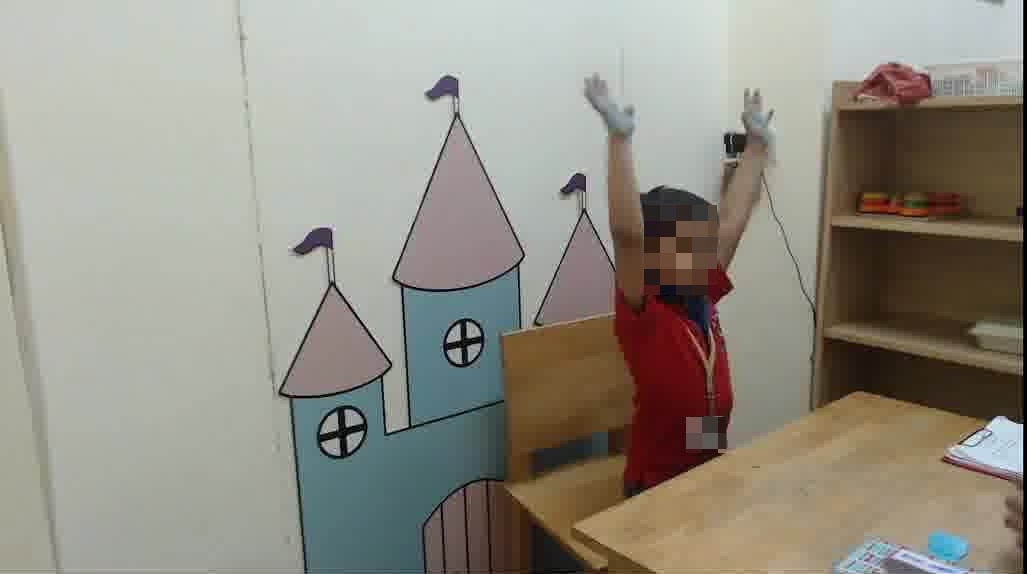}
        a) Asian subject. `Arms up' action.\label{fig:1}
    \end{minipage}
    \begin{minipage}[t]{.152\textwidth}
        \centering
        \includegraphics[width=\textwidth,height=0.44\textwidth]{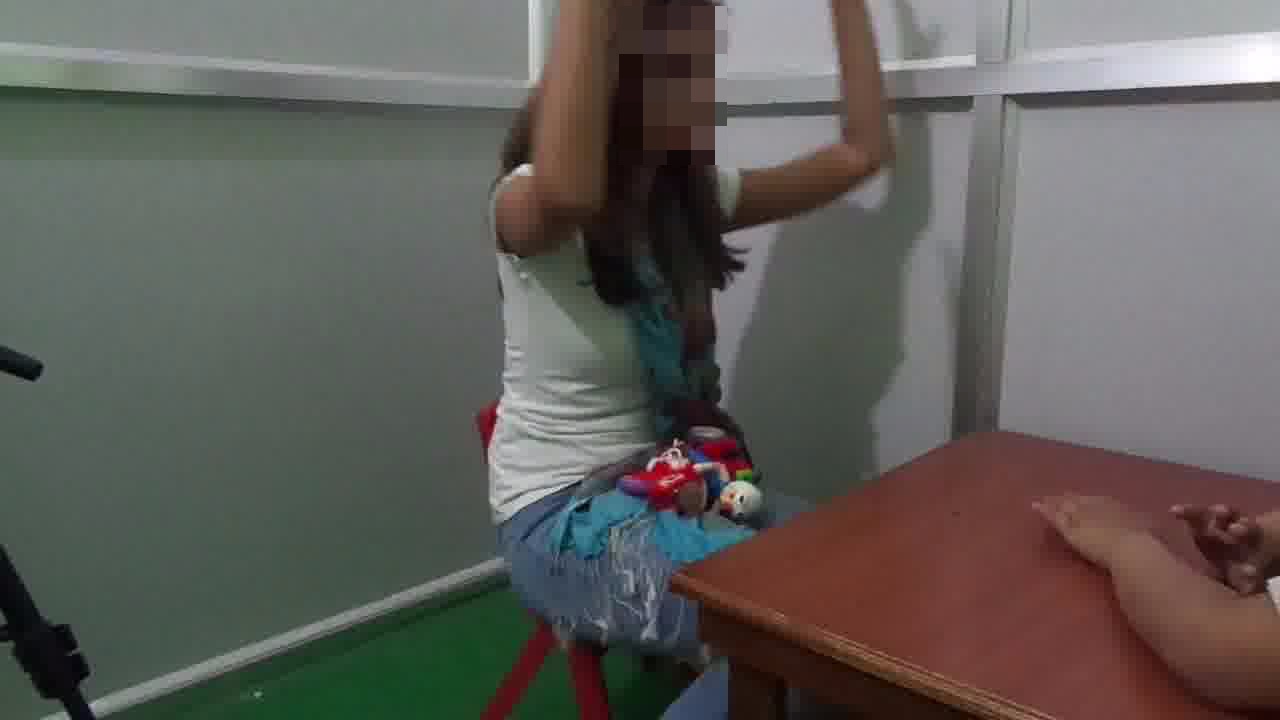}
        b) Asian clinician. Imitation instruction.\label{fig:2}
    \end{minipage}
    \begin{minipage}[t]{.152\textwidth}
        \centering
        \includegraphics[width=\textwidth,height=0.44\textwidth]{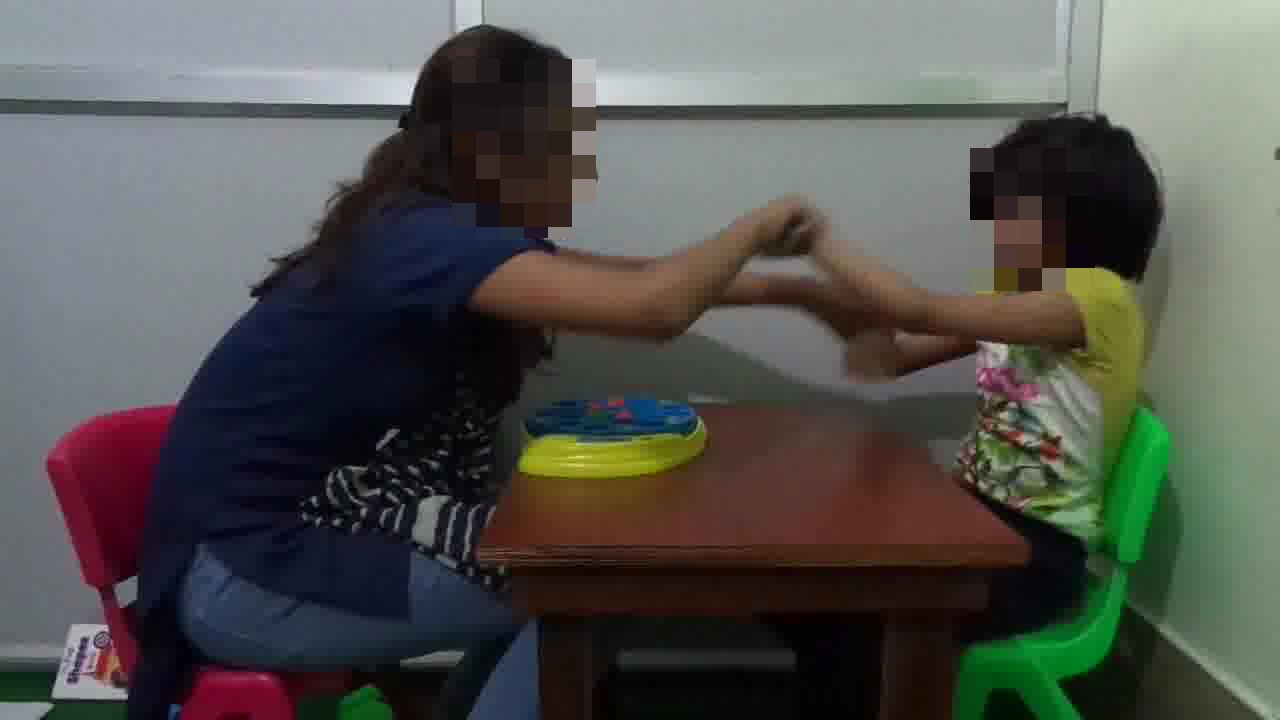}
        c) Lateral view. `Rolly polly' action. \label{fig:3}
    \end{minipage} 
    \\\\\\
    \begin{minipage}[t]{.152\textwidth}
        \centering
        \includegraphics[width=\textwidth,height=0.44\textwidth]{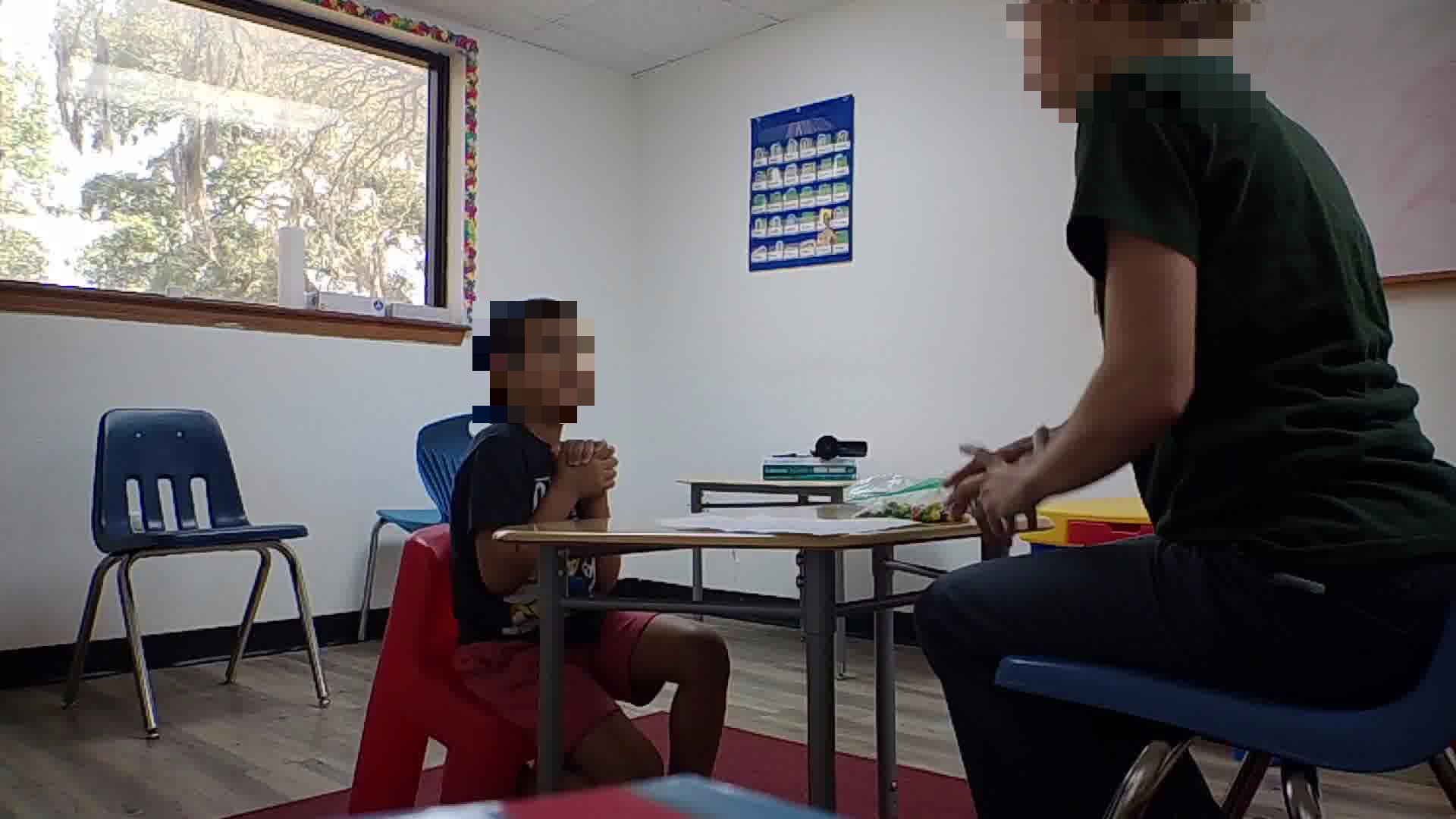}
        d) Caucasian child. `Lock hands' action.\label{fig:4}
    \end{minipage}
    \begin{minipage}[t]{.152\textwidth}
        \centering
        \includegraphics[width=\textwidth,height=0.44\textwidth]{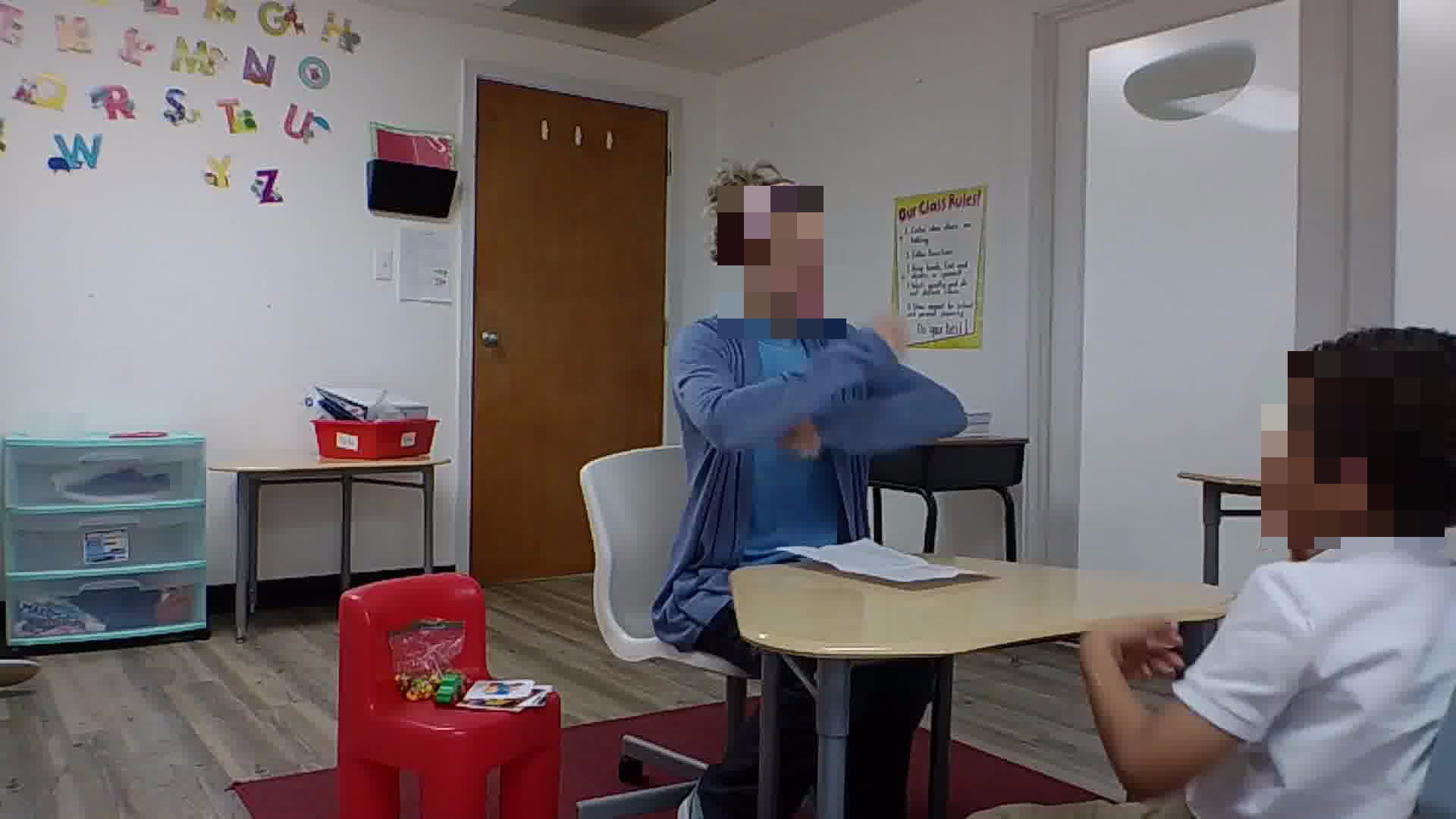}
        e) `Rolly polly'- Caucasian clinician.\label{fig:5}
    \end{minipage}
    \begin{minipage}[t]{.152\textwidth}
        \centering
        \includegraphics[width=\textwidth,height=0.44\textwidth]{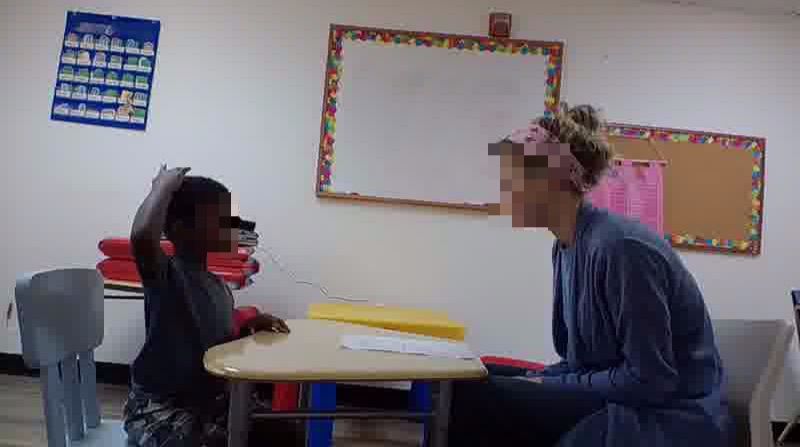}
        f) `Touch head'- Listener response.\label{fig:6}
    \end{minipage} 
    
    \caption{Training sessions of Asian and Caucasian subjects by clinicians with different camera positions. }
    \label{fig:data-des}
\end{figure}

The target Autism data consists of 37 subjects (`child' and `subject' are used interchangeably). Five subjects have Caucasian origin and the rest are Asians. Their ages range from 2-14.  
During assessment, a clinician performs the functional assessment by probing a child for age-appropriate listener response and imitation skills by invoking an instruction response and expecting a child to respond through a human action. 
We deliberately chose eight representative human action responses invoked through either listener response or imitation instruction for our experiments. Specifically, the action classes selected are - \textit{`Move the table'} and \textit{`Arms up'} for gross motor skill assessment, \textit{`Lock hands'} and \textit{`Tapping'} for fine motor skills, \textit{`Rolly polly'} for complex motor action,  \textit{`Touch nose'}, \textit{`Touch head'} and \textit{`Touch ear'} for identification of different parts of the body. We chose these particular actions since the presence of age-appropriate fine and gross motor skills demonstrate neuro-typical development of a child, as well as providing a clear picture of atypical development \citep{gowen2013motor}. 
 A total of 1481 video clips were recorded in a semi-structured environment with the clinician facing the child, and three synchronized cameras (Logitech C922 720p 60fps) were placed to record the videos. The first camera faced the clinician, the second faced the child and the third was positioned laterally to both the clinician and the child. Figure \ref{fig:data-des} depicts representative frames of some action classes during training sessions for different camera positions. The video clips were annotated by trained clinical psychologists. The response of the child for a particular stimulus is treated as a human action response classification problem. 2D-videos were recorded with 37 subjects and 12 licensed clinicians in two phases for a period of 12 months in the US and Asia across different Autism diagnostic centers. There are 148 annotated video clips and the test set contains 1333 clips. All the training and test samples are trimmed video clips containing exactly one of the eight actions of 2-5 seconds in duration. A copy of the Autism dataset can be obtained by contacting the authors.
\begin{figure*}[!htb]
\centering
    \begin{minipage}[t]{.24\linewidth}
        \centering
        \includegraphics[width=\linewidth,height=0.94\textwidth]{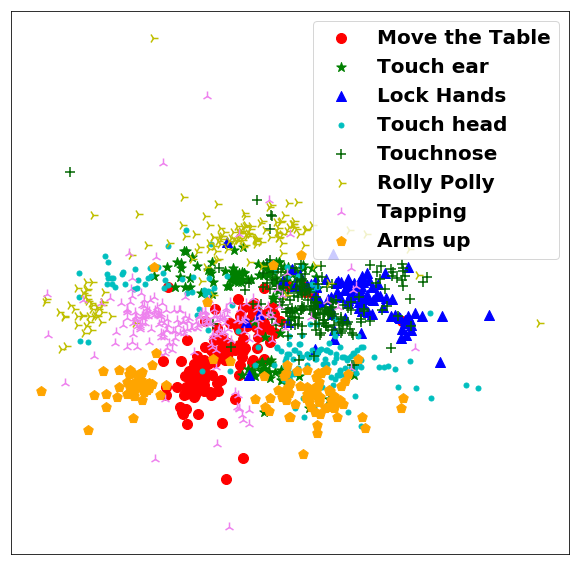}
        a) Baseline model on I3D. \label{fig:i3dtsnenodr}
    \end{minipage}
    \begin{minipage}[t]{.240\linewidth}
        \centering
        \includegraphics[width=\linewidth,height=0.94\textwidth]{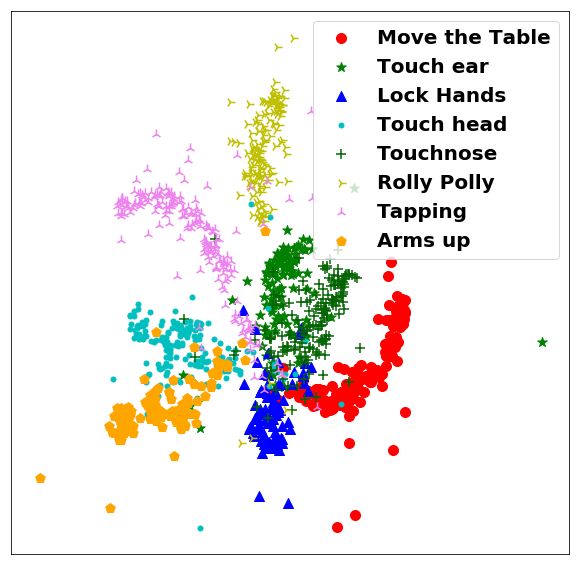}
        b) Baseline model on I3D with GWS+DR.\label{fig:i3dtsnedr}
    \end{minipage}
    \begin{minipage}[t]{.240\linewidth}
        \centering
        \includegraphics[width=\linewidth,height=0.94\textwidth]{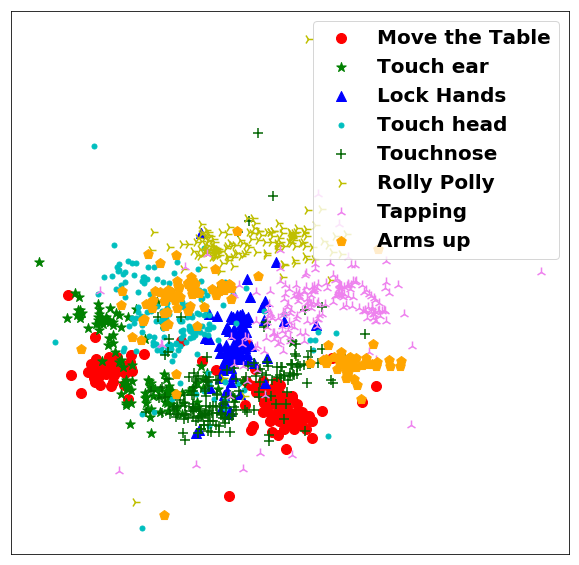}
        c) Baseline model on TSN. \label{fig:tsntsnenodr}
    \end{minipage}
    \begin{minipage}[t]{.240\linewidth}
        \centering
        \includegraphics[width=\linewidth,height=0.94\textwidth]{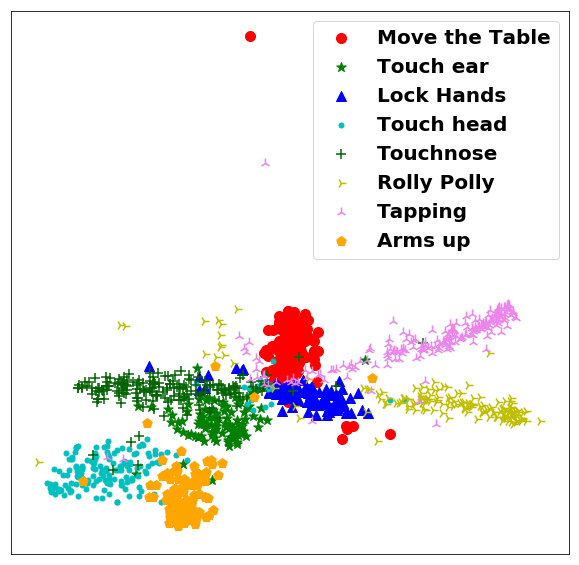}
        d) Baseline model on TSN with GWS+DR.\label{fig:tsncfdr}
    \end{minipage}
    \caption{t-SNE plots of embeddings of penultimate layers of baseline I3D and TSN Autism models with and without GWS and DR. It is clearly seen that the inter-class separability has increased and clusters are more dense after GWS and DR. }
    \label{fig:tsne}
\end{figure*}
\section{Experiments and Results}
Two large-scale publicly available human action recognition datasets namely Kinetics \citep{kay2017kinetics} and HMDB51 \citep{kuehne2011hmdb} are used as the source datasets while the data described in the previous section (termed as the Autism dataset) is the target dataset. The task is of standard 8-class classification on the target Autism data with classes as described in the previous section. We train a binary action classifier with two classes, \textit{`Response'} and \textit{`No Response'}. The \textit{`Response'} class has samples from all the 8 action classes while the \textit{`No Response'} class has samples containing random actions (different from the 8 actions in the Autism dataset) or no action at all. This classifier helps to mark the video clips with actions not belonging to one of the 8 classes or having no response. All the further experiments are carried out on video clips containing a \textit{`Response'}.
\subsection{Baselines}
For mode matching experiments, I3D is used in conjunction with Kinetics and TSN with HMDB51. 
Our baselines are I3D and TSN which are pre-trained on Kinetics and HMDB51 respectively.
Next, using the baseline model, the Kinetics and HMDB51 classes are mode matched to Autism classes. Table \ref{table:tab2} shows matched classes on both the source datasets. It is apparent from  Table \ref{table:tab2} that mode matching maps semantically similar actions from Kinetics and HMDB51 to the Autism actions, confirming the proposed hypothesis. 
\begin{table}[hbt!]
    \centering
    \resizebox{.95\columnwidth}{!}{
    \begin{tabularx}{\columnwidth}{@{\hskip 0.13in}l@{\hskip 0.03in}l@{\hskip 0.032in}XXXccc}
        \toprule
        \textbf{Autism} & \textbf{Kinetics} & \textbf{HMDB51}  \\  
        \midrule
         Move the table & Pushing cart & Push \\ 
         Touch ear & Tying necktie & Sit up \\ 
         Lock hands & Playing trombone & Shake hands \\ 
         Touch head & Blowdrying hair& Shoot ball \\ 
         Touch nose & Putting on eyeliner& Eat \\ 
         Rolly polly & Playing hand clapping games& Flic flac \\ 
         Tapping & Playing drums & Chew \\ 
         Arms up & Jumping jacks& Fall floor \\ 
         \bottomrule
    \end{tabularx}
    }
    \caption{Action classes from the source datasets (Kinetics and HMDB51)  matched to target (Autism) classes.}
    \label{table:tab2}
\end{table}
Figure \ref{fig:i3dTSN} shows the variation of test accuracy with different amounts of source data from Kinetics and HMDB51 mode matched classes augmented with Autism classes. 
Baseline models are re-trained with varying amounts of mode matched Kinetics or HMDB51 samples (equivalent to 5\%($\approx$ 75 samples) to 30\%($\approx$ 450 samples) of all the Autism samples). In all of these baseline models, when the size of augmented source data is increased and the model is re-trained, the baseline accuracy increases till the  percentage of the augmented source data is comparable (in terms of number of samples) with the Autism data. It is seen that when the source data dominates the Autism data, the accuracy drops. This is expected because when the source distribution dominates, the classifier tends to overfit on it. However, with GWS along with DR, not only the rise in test accuracy is more, the drop in accuracy after peaking is smoother as compared to GWS without DR. This implies that DR offers more tolerance towards the augmented source data which allows the performance to increase further. Our approach outperforms all baseline Autism models of I3D and TSN with comparable source samples. When we re-train the baseline Autism I3D classifier by augmenting random samples from the source (without mode matching), the test accuracy drops to 32\% thereby showing importance of GWS. 
t-SNE \citep{maaten2008visualizing} plots for the 8-dimensional embeddings from penultimate layer of I3D and TSN are obtained  for the baseline models with and without GWS and DR as shown in Figure \ref{fig:tsne}. With our approach, the inter-class separability of samples has increased while the intra-class separability has decreased so the model tends to be more confident in its predictability of Autism classes. 
\begin{figure}[hbt!]
\captionsetup{justification=raggedright}
    \begin{minipage}[t]{.495\linewidth}
        \centering
        \includegraphics[width=1.0\linewidth,height=0.57\linewidth]{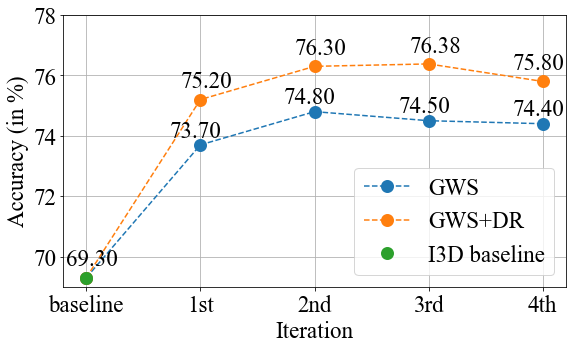}
        a) Baseline I3D with Kinetics samples.
        \label{fig:ti3ditr}
    \end{minipage}
    \begin{minipage}[t]{.495\linewidth}
        \centering
        \includegraphics[width=1.0\linewidth,height=0.57\linewidth]{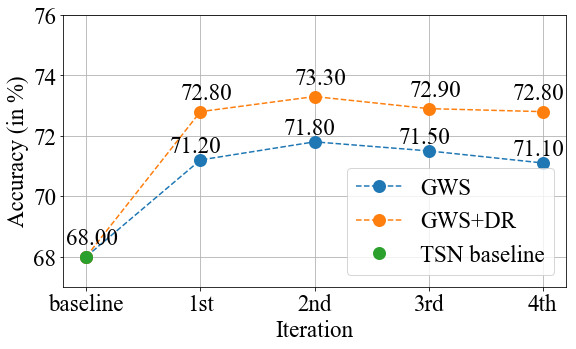}
        b) Baseline TSN with HMDB51 samples.\label{fig:tsnitr}
    \end{minipage}
    \caption{Performance on I3D and TSN with iteration over the source samples with GWS and DR. The accuracy increases through iteration by augmenting with newer mode matches samples in every iteration although overfitting on source data occurs from \nth{3} iteration.}
\label{fig:itrTSNi3d}
\end{figure}
\begin{table}[hbt!]
\centering
  \resizebox{.97\columnwidth}{!}{
  \begin{tabular}{lcccc}
    \toprule
        \textbf{Model} & \textbf{\vtop{\hbox{\strut Asian}}} 
        & \textbf{\vtop{\hbox{\strut Caucasian}}}
        & \textbf{\vtop{\hbox{\strut $\leq$ 5 yrs.}}} & \textbf{\vtop{\hbox{\strut $>$ 5 yrs.}}} 
        \\
    \midrule
    I3D&58.2\%&34.6\%&55.1\%&41.2\%\\
    TSN&53.7\%&30.2\%&51.7\%&38.1\%\\
    I3D+GWS+DR&\textbf{63.7\%}&\textbf{40.2\%}&\textbf{59.8\%}&\textbf{46.5\%}\\
    TSN+GWS+DR&\textbf{57.0\%}&\textbf{33.8\%}&\textbf{55.4\%}&\textbf{40.9\%}\\
    \bottomrule
  \end{tabular}
  }
  \caption{Performance of GWS and DR with a specific bias in the Autism training dataset. It is seen that in all the cases, the proposed approach offers better performance over the baselines.}
  \label{table:bias}
\end{table}
\begin{table*}[hbt!]
\centering
  \scalebox{0.92}{
  \begin{tabular}{lcccccc}
    \toprule
        \textbf{Model} & \textbf{\vtop{\hbox{\strut Second best}\hbox{\strut modes}}} 
        & \textbf{\vtop{\hbox{\strut Flow + RGB}\hbox{\strut stream}}}
        & \textbf{\vtop{\hbox{\strut Handpicked}\hbox{\strut modes}}} & \textbf{\vtop{\hbox{\strut Combined}\hbox{\strut modes}}} & \textbf{I3D\textrightarrow TSN} &  \textbf{TSN\textrightarrow I3D}
        \\
    \midrule
    I3D+GWS&\textbf{73.5\%}&67.5\%&\textbf{74.2\%}&\textbf{73.8\%}&-&\textbf{73.2\%}\\
    TSN+GWS&\textbf{71.4\%}&65.8\%&\textbf{71.9\%}&\textbf{71.2\%}&\textbf{71.7\%}&-\\
    I3D+GWS+DR&\textbf{74.4\%}&68.1\%&\textbf{75.3\%}&\textbf{74.8\%}&-&\textbf{74.4\%}\\
    TSN+GWS+DR&\textbf{72.4\%}&66.7\%&\textbf{72.8\%}&\textbf{72.7\%}&\textbf{73.1\%}&-\\
    \bottomrule
  \end{tabular}
  }
  \caption{GWS and DR under different settings on baselines (69\% for I3D and 68\% for TSN) - It is seen that (a) GWS or GWS+DR with second best modes too leads in better performance (b) re-training the RGB stream is detrimental since there is no similarity in the RGB space (c) GWS or GWS+DR with hand-picked modes also results in improvement in accuracy (d) re-training with samples from matched modes classes from different datasets results in performance enhancement (e) \& (f) Cross neural architecture GWS and DR - Increase in accuracy from the baselines indicate that similar actions in the optical flow space retain their meaning irrespective of the neural architectures.}
  \label{table:diffsettings}
\end{table*}
\begin{table*}[hbt!]
\centering
  \scalebox{0.92}{
  \begin{tabular}{lcccccc}
    \toprule
        \textbf{Model} & \textbf{\vtop{\hbox{\strut Autism}\hbox{\strut Dataset}}} 
        & \textbf{\vtop{\hbox{\strut SSBD}\hbox{\strut  }}}
       
        \\
    \midrule
    TSN  (Pretrained with HMDB51) \citep{wang2016temporal}&68.0\%&87.4\%\\
    I3D   (Pretrained with Kinetics) \citep{carreira2017quo}&69.3\%&91.2\%\\
    ECO (Pretrained with Kinetics) \citep{zolfaghari2018eco}&61.4\%&80.1\%\\
    TSM (Pretrained with Kinetics) \citep{lin2019tsm}&69.8\%&90.5\%\\
    R(2+1)D (Pretrained with IG-65M) \citep{ghadiyaram2019large}&68.4\%&88.3\%\\
     \midrule
    TSN+DR &70.1\%&89.2\%\\
    I3D+DR &71.3\%&92.8\%\\
    TSN+GWS &71.6\%&90.3\%\\
    I3D+GWS &74.3\%&93.6\%\\
    TSN+GWS+DR &72.5\%&91.4\%\\
    I3D+GWS+DR &\textbf{75.1}\%&\textbf{95.7}\%\\
    \bottomrule
  \end{tabular}
  }
  \caption{Comparison with state-of-the-art action recognition models on Autism dataset and SSBD.}
  \label{table:comparison}
\end{table*}
In the next set of experiments,  we iteratively re-train the target model with GWS and DR.  That is, in every new iteration we discard the mode matched source samples from the previous iteration keeping the number of samples similar in every iteration (equivalent to the number of Autism training samples) and re-train the target model with new set of samples from the matched modes classes. Figure \ref{fig:itrTSNi3d} shows the variation in accuracy when the new source samples are augmented with Autism data in every iteration. In Figure 6a, the test accuracy increases from the baseline for initial iterations with GWS and DR. With GWS only (I3D+GWS), the accuracy drops after the \nth{2} iteration. If GWS is applied with DR on the baseline (I3D+GWS+DR), it continues to increase even after the \nth{2} iteration but starts to dip from \nth{3} iteration on-wards which can be ascribed to overfitting on the source data.  As with the previous cases, we observe more tolerance of DR with newer samples as compared to GWS. Newer samples are accepted with lesser surprise in DR which enhances the generalizability and performance. Similar behavior is observed with TSN as shown in Figure 6b. 
\subsection{Bias in the Training Data}
Table \ref{table:bias} shows the results  for the proposed method under different kinds of dataset biases. The results in the second column in this table are accuracies when the training data has only Asian subjects. Third column are test accuracies when the training data has Caucasian subjects. In the fourth column, the training data has subjects that are 5 years or below. The last column are test accuracies when the training data has subjects above 5 years in age. It is observed that even if the dataset has a bias, our approach performs better than the baselines.
\subsection{GWS and DR under Different Settings}
Table \ref{table:diffsettings} shows test accuracy scores on baseline Autism model under six different settings. We sampled examples from second best modes to augment with Autism data using GWS and DR. The results in the second column are the test accuracy scores using second best modes. It can be seen that the performance is better than the baseline model albeit less than the performance of model with the best modes shown in Table \ref{table:tab2}.  It is however apparent that even second best modes preserve the closeness in the optical flow space. All the experiments are executed by re-training optical flow stream of the baseline models. Besides optical flow, if we re-train the RGB stream as well, it is seen that  the performance of the classifiers deteriorates as shown in third column of Table \ref{table:diffsettings}. This ascertains the fact that the notion of closeness is valid only for optical flow samples. Next, we handpicked similar action classes (like \textit{`Washing hands'} in Kinetics matched to \textit{`Lock hands'} in Autism data, etc.) from Kinetics and HMDB51 datasets and applied our approach on the baseline models. Fourth column of Table \ref{table:diffsettings} records test accuracy scores with these handpicked classes (or modes). The results are comparable to our approach where we employ baseline models to find similar classes or actions using mode matching. Hence, irrespective of the metric used to find similarity (either using human intelligence or mode matching), the performance is much better than the baselines when they are augmented and re-trained with similar classes (GWS) with or without DR. Next, we combined corresponding best mode matched classes from Kinetics and HMDB51 (like \textit{`Pushing cart'} from Kinetics is combined with \textit{`Push'} from HMDB51, etc.) and used these samples for GWS and DR as shown in the fifth column of Table \ref{table:diffsettings}. The performance is comparable with our approach when the modes are not combined which implies that the idea of similarity in optical flow space is preserved even if different datasets are merged. Fifth and sixth columns report test accuracies for the case when the modes matched using one architecture is used to augment the classifier built on another architecture. Fifth column of Table \ref{table:diffsettings} are the test accuracy scores when baseline TSN is re-trained with augmented Kinetics modes extracted from mode matching on I3D. Similarly, sixth column has test accuracy scores when baseline I3D is re-trained with HMDB51 modes extracted from TSN. The scores are still better than the baseline Autism model (69\% for I3D and 68\% for TSN) which indicates that similar actions in the optical flow space retain their meaning irrespective of the neural architectures used as a backbone.
\subsection{Comparisons with State-of-the-art}
Table \ref{table:comparison} reports test accuracy for GWS and DR on baseline I3D and TSN.
It is consistently observed that with our approach (GWS or DR or GWS+DR), the performance is better than the baseline I3D and TSN and state-of-the-art action recognition models using Autism dataset and SSBD. SSBD is a public dataset having 75 examples and three action classes namely \textit{`arm flapping'}, \textit{`head banging'} and \textit{`spinning'} that are used in Autism diagnosis. Additionally, it is seen that DR offers additional accuracy benefits over GWS. 
\section{Conclusion}
In this paper, we proposed a method for improving the generalization abilities of a classifier designed for human action recognition trained on scarce data. Specifically, leveraging the semantic similarities of the action classes in the optical flow space, we proposed a generic method called Guided Weak Supervision (GWS) to augment and re-train a classifier on the target data with samples from a large-scale annotated dataset, along with a novel loss function termed as Directional Regularization (DR) that would result in performance enhancement. We demonstrated the efficacy of the proposed method for screening, diagnosis and behavioral treatment for ASD by treating imitation, listener response and motor skills as action classification tasks.  With the proposed framework, we can integrate and automate complete value chain of screening to treatment for children with ASD by capturing behavioral treatment progress data temporally and remotely. The other future direction is to employ the techniques of GWS and DR for generalized supervised learning tasks beyond the proposed use case of Autism.
\bibliographystyle{aaai}
\bibliography{mybibliography}

\begin{thebibliography}{}

\bibitem[\protect\citeauthoryear{Carreira and
  Zisserman}{2017}]{carreira2017quo}
Carreira, J., and Zisserman, A.
\newblock 2017.
\newblock Quo vadis, action recognition? a new model and the kinetics dataset.
\newblock In {\em proceedings of the IEEE Conference on Computer Vision and
  Pattern Recognition},  6299--6308.

\bibitem[\protect\citeauthoryear{CDC}{2019}]{CDC}
2019.
\newblock Data \& statistics on autism spectrum disorder.
\newblock \url{https://www.cdc.gov/ncbddd/autism/data.html}.
\newblock Accessed: 2019-06-30.

\bibitem[\protect\citeauthoryear{Donahue \bgroup et al\mbox.\egroup
  }{2015}]{donahue2015long}
Donahue, J.; Anne~Hendricks, L.; Guadarrama, S.; Rohrbach, M.; Venugopalan, S.;
  Saenko, K.; and Darrell, T.
\newblock 2015.
\newblock Long-term recurrent convolutional networks for visual recognition and
  description.
\newblock In {\em Proceedings of the IEEE conference on computer vision and
  pattern recognition},  2625--2634.

\bibitem[\protect\citeauthoryear{Estes \bgroup et al\mbox.\egroup
  }{2015}]{estes2015long}
Estes, A.; Munson, J.; Rogers, S.~J.; Greenson, J.; Winter, J.; and Dawson, G.
\newblock 2015.
\newblock Long-term outcomes of early intervention in 6-year-old children with
  autism spectrum disorder.
\newblock {\em Journal of the American Academy of Child \& Adolescent
  Psychiatry} 54(7):580--587.

\bibitem[\protect\citeauthoryear{Ghadiyaram, Tran, and
  Mahajan}{2019}]{ghadiyaram2019large}
Ghadiyaram, D.; Tran, D.; and Mahajan, D.
\newblock 2019.
\newblock Large-scale weakly-supervised pre-training for video action
  recognition.
\newblock In {\em Proceedings of the IEEE Conference on Computer Vision and
  Pattern Recognition},  12046--12055.

\bibitem[\protect\citeauthoryear{Gowen and Hamilton}{2013}]{gowen2013motor}
Gowen, E., and Hamilton, A.
\newblock 2013.
\newblock Motor abilities in autism: a review using a computational context.
\newblock {\em Journal of autism and developmental disorders} 43(2):323--344.

\bibitem[\protect\citeauthoryear{Horn and Schunck}{1981}]{horn1981determining}
Horn, B.~K., and Schunck, B.~G.
\newblock 1981.
\newblock Determining optical flow.
\newblock {\em Artificial intelligence} 17(1-3):185--203.

\bibitem[\protect\citeauthoryear{Kay \bgroup et al\mbox.\egroup
  }{2017}]{kay2017kinetics}
Kay, W.; Carreira, J.; Simonyan, K.; Zhang, B.; Hillier, C.; Vijayanarasimhan,
  S.; Viola, F.; Green, T.; Back, T.; Natsev, P.; et~al.
\newblock 2017.
\newblock The kinetics human action video dataset.
\newblock {\em arXiv preprint arXiv:1705.06950}.

\bibitem[\protect\citeauthoryear{Kuehne \bgroup et al\mbox.\egroup
  }{2011}]{kuehne2011hmdb}
Kuehne, H.; Jhuang, H.; Garrote, E.; Poggio, T.; and Serre, T.
\newblock 2011.
\newblock Hmdb: a large video database for human motion recognition.
\newblock In {\em 2011 International Conference on Computer Vision},
  2556--2563.
\newblock IEEE.

\bibitem[\protect\citeauthoryear{Lin, Gan, and Han}{2019}]{lin2019tsm}
Lin, J.; Gan, C.; and Han, S.
\newblock 2019.
\newblock Tsm: Temporal shift module for efficient video understanding.
\newblock In {\em Proceedings of the IEEE International Conference on Computer
  Vision},  7083--7093.

\bibitem[\protect\citeauthoryear{Maaten and
  Hinton}{2008}]{maaten2008visualizing}
Maaten, L. v.~d., and Hinton, G.
\newblock 2008.
\newblock Visualizing data using t-sne.
\newblock {\em Journal of machine learning research} 9(Nov):2579--2605.

\bibitem[\protect\citeauthoryear{Mandal \bgroup et al\mbox.\egroup
  }{2019}]{Mandal_2019_CVPR}
Mandal, D.; Narayan, S.; Dwivedi, S.~K.; Gupta, V.; Ahmed, S.; Khan, F.~S.; and
  Shao, L.
\newblock 2019.
\newblock Out-of-distribution detection for generalized zero-shot action
  recognition.
\newblock In {\em The IEEE Conference on Computer Vision and Pattern
  Recognition (CVPR)}.

\bibitem[\protect\citeauthoryear{Rapin}{1991}]{rapin1991autistic}
Rapin, I.
\newblock 1991.
\newblock Autistic children: Diagnosis and clinical features.
\newblock {\em Pediatrics} 87(5):751--760.

\bibitem[\protect\citeauthoryear{Simonyan and
  Zisserman}{2014}]{simonyan2014two}
Simonyan, K., and Zisserman, A.
\newblock 2014.
\newblock Two-stream convolutional networks for action recognition in videos.
\newblock In {\em Advances in neural information processing systems},
  568--576.

\bibitem[\protect\citeauthoryear{Soomro, Zamir, and
  Shah}{2012}]{soomro2012ucf101}
Soomro, K.; Zamir, A.~R.; and Shah, M.
\newblock 2012.
\newblock Ucf101: A dataset of 101 human actions classes from videos in the
  wild.
\newblock {\em arXiv preprint arXiv:1212.0402}.

\bibitem[\protect\citeauthoryear{Sundar~Rajagopalan, Dhall, and
  Goecke}{2013}]{Rajagopalan_2013_ICCV_Workshops}
Sundar~Rajagopalan, S.; Dhall, A.; and Goecke, R.
\newblock 2013.
\newblock Self-stimulatory behaviours in the wild for autism diagnosis.
\newblock In {\em The IEEE International Conference on Computer Vision (ICCV)
  Workshops}.

\bibitem[\protect\citeauthoryear{Tran \bgroup et al\mbox.\egroup
  }{2015}]{tran2015learning}
Tran, D.; Bourdev, L.; Fergus, R.; Torresani, L.; and Paluri, M.
\newblock 2015.
\newblock Learning spatiotemporal features with 3d convolutional networks.
\newblock In {\em Proceedings of the IEEE international conference on computer
  vision},  4489--4497.

\bibitem[\protect\citeauthoryear{Wang \bgroup et al\mbox.\egroup
  }{2016}]{wang2016temporal}
Wang, L.; Xiong, Y.; Wang, Z.; Qiao, Y.; Lin, D.; Tang, X.; and Van~Gool, L.
\newblock 2016.
\newblock Temporal segment networks: Towards good practices for deep action
  recognition.
\newblock In {\em European conference on computer vision},  20--36.
\newblock Springer.

\bibitem[\protect\citeauthoryear{Wang, Zhou, and Qiao}{2018}]{Wang_2018_CVPR}
Wang, Y.; Zhou, L.; and Qiao, Y.
\newblock 2018.
\newblock Temporal hallucinating for action recognition with few still images.
\newblock In {\em The IEEE Conference on Computer Vision and Pattern
  Recognition (CVPR)}.

\bibitem[\protect\citeauthoryear{Yang, He, and Porikli}{2018}]{Yang_2018_CVPR}
Yang, H.; He, X.; and Porikli, F.
\newblock 2018.
\newblock One-shot action localization by learning sequence matching network.
\newblock In {\em The IEEE Conference on Computer Vision and Pattern
  Recognition (CVPR)}.

\bibitem[\protect\citeauthoryear{Zach, Pock, and
  Bischof}{2007}]{zach2007duality}
Zach, C.; Pock, T.; and Bischof, H.
\newblock 2007.
\newblock A duality based approach for realtime tv-l 1 optical flow.
\newblock In {\em Joint pattern recognition symposium},  214--223.
\newblock Springer.

\bibitem[\protect\citeauthoryear{Zhu and Yang}{2018}]{Zhu_2018_ECCV}
Zhu, L., and Yang, Y.
\newblock 2018.
\newblock Compound memory networks for few-shot video classification.
\newblock In {\em The European Conference on Computer Vision (ECCV)}.

\bibitem[\protect\citeauthoryear{Zolfaghari, Singh, and
  Brox}{2018}]{zolfaghari2018eco}
Zolfaghari, M.; Singh, K.; and Brox, T.
\newblock 2018.
\newblock Eco: Efficient convolutional network for online video understanding.
\newblock In {\em Proceedings of the European Conference on Computer Vision
  (ECCV)},  695--712.

\end{thebibliography}
\end{document}